\documentclass[10pt,twocolumn,letterpaper]{article}

\usepackage{cvpr}              

\usepackage{amsmath}
\usepackage{algorithm}
\usepackage{algpseudocode}
\usepackage{graphicx}
\usepackage{subcaption}
\usepackage{caption}

\usepackage{booktabs} 

\usepackage[dvipsnames]{xcolor}

\definecolor{cvprblue}{rgb}{0.21,0.49,0.74}
\usepackage[pagebackref,breaklinks,colorlinks,citecolor=cvprblue]{hyperref}

\def\paperID{21} 
\def\confName{CVPR}
\def\confYear{2024}

\title{Parameter-Efficient Active Learning for Foundational models}

\author{Athmanarayanan Lakshmi Narayanan\\
Intel labs, USA\\
Santa Clara,CA,USA\\
{\tt\small athma.lakshmi.narayanan@intel.com}
\and
Ranganath Krishnan\\
Intel labs, USA\\
Hillsboro,OR,USA\\
{\tt\small ranganath.krishnan@intel.com}
\and
Amrutha Machireddy\\
Intel labs, India\\
Bangalore,Karnataka,India\\
{\tt\small amrutha.machireddy@intel.com}
\and
Mahesh Subedar\\
Intel labs, USA\\
Hillsboro,OR,USA\\
{\tt\small mahesh.subedar@intel.com}
}

\begin{document}
\maketitle
\begin{abstract}
Foundational vision transformer models have shown impressive few shot performance on many vision tasks. This research presents a novel investigation into the application of parameter efficient fine-tuning methods within an active learning (AL) framework, to advance the sampling selection process in extremely budget constrained classification tasks. The focus on image datasets, known for their out-of-distribution characteristics, adds a layer of complexity and relevance to our study. Through a detailed evaluation, we illustrate the improved AL performance on these challenging datasets, highlighting the strategic advantage of merging parameter efficient fine tuning methods with foundation models. This contributes to the broader discourse on optimizing AL strategies, presenting a promising avenue for future exploration in leveraging foundation models for efficient and effective data annotation in specialized domains.
\end{abstract}    
\section{Introduction}
\label{sec:intro}
The emergence of foundation models~\cite{clip, dinov2, sam} has profoundly advanced the field of machine learning, delivering unmatched performance across a wide range of applications. Their remarkable scalability and adaptability have set high standards in efficiency and effectiveness, marking a new era in the field of AI. Vision Transformer (ViT) based foundation models have demonstrated remarkable proficiency in zero-shot and few-shot transfer learning~\cite{park2024pre} with their ability to generalize from minimal examples through the self-attention mechanism. However, identifying the most informative minimal data samples for few-shot transfer learning is challenging, as there is a significant cost and time associated with data annotation, which is critical for ensuring that the foundation models are trained on high-quality and representative examples. In this context, active learning~\cite{settles2009active, ren2021survey} emerges as a strategic approach to enhance the efficiency of data annotation process. Active learning involves selecting the most valuable data instances for labeling, aiming to boost model accuracy with minimal data.

This work explores the effective utilization of active learning with foundation models, investigating the potential of data-efficient and parameter-efficient few shot transfer learning. Though active learning is a well-studied and ongoing research area~\cite{ren2021survey}, the utilization of active sample selection techniques with vision transformer backbones remains an underexplored area in the age of foundation models. We investigate how active learning's selective data querying can be harmonized with the broad learning capacity of foundation models, focusing on identifying and labeling the most impact data samples. Through extensive empirical evaluation on large-scale image classification datasets, we illustrate how active learning can effectively decrease the volume of necessary training data for foundation models, thereby optimizing transfer learning resource allocation.


\noindent The key contributions of this work include:
\begin{itemize}
\item We introduce Parameter-Efficient Active Learning (PEAL) for foundation models to enable effective transfer learning from most informative data samples.
\item The proposed approach enables effective use of feature-embedding based sample selection strategies in active learning settings with foundation models.
\item We demonstrate PEAL improves the transfer learning performance and efficiency with foundation models, as compared to linear probing. We empirically show the effectiveness of PEAL for both uncertainty-based and diversity-based sample selection methods with extensive experiments on large-scale image-classification datasets utilizing DINOv2~\cite{dinov2} as the foundation backbone.

\end{itemize}

\section{Related Work}
\label{sec:formatting}
Recent advances in active learning (AL)~\cite{ren2021survey} have demonstrated significant improvements in data annotation and model learning efficiency. AL methods can be broadly classified into two major sampling strategies, \textit{uncertainty} \cite{joshi2009multi,ranganathan2017deep,krishnan2021robust,yoo2019learning} based and \textit{diversity} based sampling~\cite{sener2017active,cho2023querying,lin2017active}. AL methods have also been studied in the context of large language transformer models~\cite{jukic2023parameter,zhang2024star} and vision-language models~\cite{bang2023active,Parvaneh_2022_CVPR}. However, the impact of foundation vision models on AL remains under explored. A recent work \cite{gupte2024revisiting}, which is more closely related to this research, investigates the use of vision foundation models in an active learning context through linear probing. However, linear probing with frozen features from the backbone limits the application of feature-embedding and diversity-based sample selection strategies within the AL framework. With parameter-efficient active learning, our work facilitates the use of feature-embedding and diversity-based active sampling strategies opening up the research scope to explore advanced active learning methods with foundation models.


\section{Methodology}

In this section, we describe our methodology for improving data annotation and model learning efficiency for vision transformer foundation models through active learning sample selection process for data-efficient transfer learning. 

\subsection{Linear Probing}
\label{sec:linprob}
 Linear probing \cite{dinov2,gupte2024revisiting} is a commonly used technique to leverage the representational power of pre-trained transformer model without the need for extensive fine-tuning. This involves training a simple linear classifier on top of the frozen features extracted from the pre-trained transformer backbone. The benefit is the minimal training overhead, eliminating the need to train the full set of backbone features that can constitute millions of parameters. 


\subsection{Parameter efficient fine tuning}
We introduce parameter-efficient active learning (PEAL) by leveraging the Low-Rank Adaptation (LoRa)~\cite{hu2021lora} for transfer learning in a data-efficient manner, enabling the model to learn from limited data by selecting the most informative samples. LoRa is a training approach that keeps the pre-trained model weights unchanged and adds trainable matrices that break down the ranks within each transformer architecture layer. This method greatly reduces the number of parameters to be trained for tasks that follow, providing a more efficient option than training all the model's weights.

We add low-rank weight matrices that are injected to the \textit{QKV} components of each attention layer in the transformer model. For DINOv2~\cite{dinov2}, this change increases the total number of trainable parameters by only 0.03\%, aside from the classifier mentioned in linear probing. This enhances the model's adaptability without significantly increasing the number of parameters, which supports our goal of parameter-efficient and effective active learning.

\subsection{Selection Strategies}
Active learning strategies are pivotal for efficiently utilizing data annotation resources. 
There are various uncertainty-based and diversity-based sample selection strategies~\cite{ren2021survey}. To assess the effectiveness of Parameter-Efficient Active Learning (PEAL) and linear probing within the active learning paradigm, we employ one method from each category: Entropy, representing \textit{uncertainty-based} sampling, and Featdist, exemplifying \textit{diversity-based} sampling. This study presents an approach that is orthogonal to, yet compatible with any existing AL sample selection strategies, offering a novel perspective on optimizing the transfer learning process with vision transformer foundation models.
\subsubsection{Uncertainty-based sampling}
We use entropy~\cite{shannonentropy} as a measure of uncertainty in the active learning process. Samples with high entropy are considered as model unknowns, and hence, potentially more informative for the model to learn from. 
$$ H(X) = -\sum_{i=1}^{n} p(x_i) \log p(x_i) $$
where, $H(X)$ is the notation for the entropy of the data sample $X$ and $p(x_i)$ represents the probability of the sample $X$ belonging to $i^{th}$ category. We select the samples with the largest entropy based on the annotation budget.
\subsubsection{Diversity-based sampling}
Feature-embedding methods such as feature distance (Featdist) measures are essential for the diversity of selected samples. With class-balanced sampling~\cite{krishnan2021robust}, we select diverse set of samples per class to enhance the representation of the chosen samples. Diversity-based sampling strategies measure sample diversity by their distance in the feature-embedding space \cite{sener2017active}. However, with linear probing restricted to frozen transformer features, distance measures become less effective. In active learning settings, to facilitate feature distance sample selection methods, the feature-embedding space must be updated to reflect the model's learning from newly annotated data. Since finetuning the entire transformer backbone is computationally intensive, we introduce parameter-efficient active learning to enable feature-embedding based AL with foundation models.

Computing the pairwise distance between each labeled and unlabeled sample poses computational challenge in resource-constrained environments. 
We compute the pairwise distances efficiently  with much lesser amount of data using the Faiss library \cite{douze2024faiss}.
Algorithm~\ref{alg:alg1} shows computing the Euclidean distance~(L2 norm) between the unlabeled and labelled samples. For each class in the labeled dataset, a dictionary is constructed based on the extracted transformer features. This dictionary has dimensions $N_c \times f$, where $N_c$ represents the number of labeled samples for  the class $c$, and $f$ denotes the feature size. Subsequently, for each unlabelled sample we  compute the model prediction and the features. We index the appropriate dictionary in $D_{c}$ using the class prediction and compute the pairwise distance between the unlabelled sample and each labelled entry in the dictionary.  We select equal number of samples for each class (class-balanced) that have the largest distance scores based on the available data annotation budget. 

\begin{algorithm}
\caption{Efficient Class-wise Sample Distance}
\label{alg:alg1}
\begin{algorithmic}[1]
\For{each labeled $l_i$}
    \State $D_c \gets \text{FaissIndex}(f(l_i))$ \Comment{$c$: class of $l_i$, $f(x)$ is the model}
\EndFor
\For{each unlabeled $u_j$}
    \State Identify class $c$ for $u_j$
    \State Define $d_{\text{max}}$ based on:
        \State $d_{\text{max}} \gets \max(\|f(u_j) - x_k\|_2), \forall x_k \in D_c$
\EndFor
\State Select top $B$ values from $d_{\text{max}}$ based on budget.
\end{algorithmic}
\end{algorithm}
\section{Experiments and Discussion}
\label{sec:Results}


\subsection{Model Architecture and Training parameters}
We use DINOv2~\cite{dinov2} with ViT-g/14~\cite{zhai2022scaling} transformer backbone as the foundation model in our experiments. However, our methodology is compatible with any vision transformer model. In the case of linear probing, a single linear layer that acts on the embedding size of $1536$ key features is added after batch normalization and  dropout ($p=0.5$) respectively. For PEAL, we inject LoRa adapters in all the \textit{QKV} attention layers. This involves only 0.03\% of the DINOv2 model parameters trainable along with the linear layer. We use  batch size of $64$, learning rate = $1 \times 10^{-3}$ and weight decay = $1 \times 10^{-2}$. We train the classifier layers and low rank layers layers  for $50$ epochs using Adam optimizer. We train the model for far fewer epochs compared to \cite{gupte2024revisiting} as we use a transfer learning setup and pick the best model for sample selection. For the LoRa, we use rank ~$=16$, alpha~$=16$ and $\text{LoRa}_{\text{dropout}} = 0.1$. The learning rate is scheduled to drop at AL cycle $5$ and $8$ with a $gamma=0.1$ to facilitate improved convergence for PEAL methods.

\begin{figure}[t!]
    \centering
    
        \centering
    \includegraphics[width=0.35\textwidth,trim=0 0 0 0in, clip]{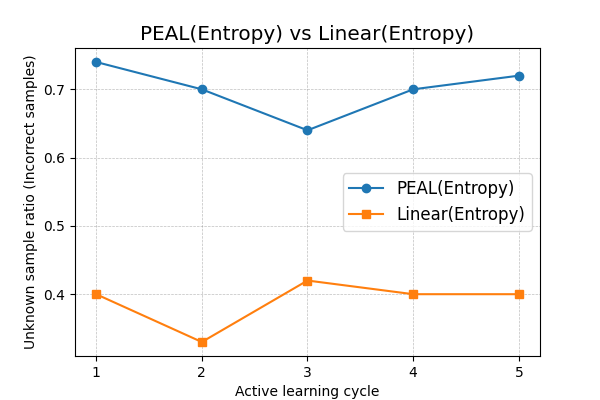}
    
    \caption{
    Comparing Linear probing and PEAL on unknown sample selection ratio (higher is better) at initial active learning cycles. The selection strategy needs to select more incorrect samples (model unknowns) than correct samples (model knowns) for effective data annotation and model learning.}
    \label{fig:histplots}
\end{figure}

\subsection{Datasets}

To evaluate the effective and efficient transfer learning with foundation model under active learning settings, we perform experiments with datasets from distinct domains that differ significantly from the general-purpose images used to pre-train the DINOv2 foundation model. Specifically, we have chosen datasets from the field of medical imaging including Histology~\cite{kather2016multi} and APTOS 2019~\cite{khalifa2019deep}, as well as satellite and aerial imagery with EuroSAT~\cite{helber2019eurosat}. The Histology dataset involves classifying textures in colorectal cancer to one of the eight categories. For experiments with Histology, we start with a pool of 4000 unlabeled images and annotation budget is set to 500 with 50 images selected for annotation iteratively every AL cycle. At every AL cycle, the model is evaluated on the 1000 test set images. The APTOS dataset contains retinal fundus images involving classification of diabetic retinopathy among five categories. For APTOS experiments, we started with 2929 unlabeled images and an annotation budget of 300, selecting 30 images per AL cycle. At every AL cycle, the model is evaluated on the test data containing 733 images. The EuroSAT dataset comprises of geo-referenced satellite images across ten categories.  For EuroSAT, we start with 24,300 unlabelled images, the annotation budget and sampling size is set to 500 and 50 images respectively. At every AL cycle, the model is evaluated on the 2700 test set images. All images are resized to $224 \times 224$ to be compatible with DINOv2 model.

\begin{figure*}[ht]
    \begin{subfigure}{0.33\linewidth}
\includegraphics[width=\textwidth, trim=0 0 0 0, clip]{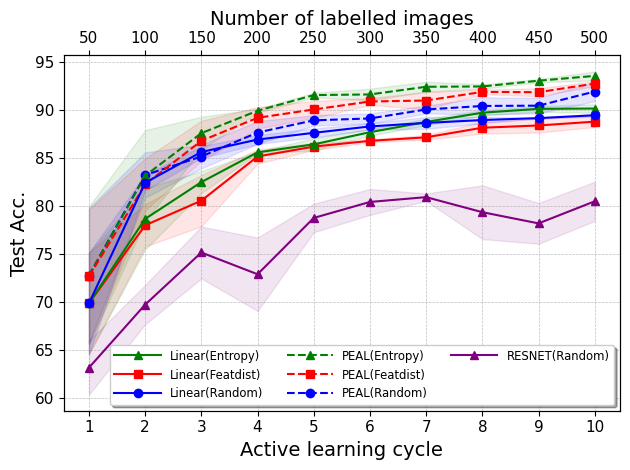}
    \caption{Histology}
    \label{fig:histology}
  \end{subfigure} 
  \begin{subfigure}{0.33\linewidth}
\includegraphics[width=\textwidth, trim=0 0 0 0, clip]{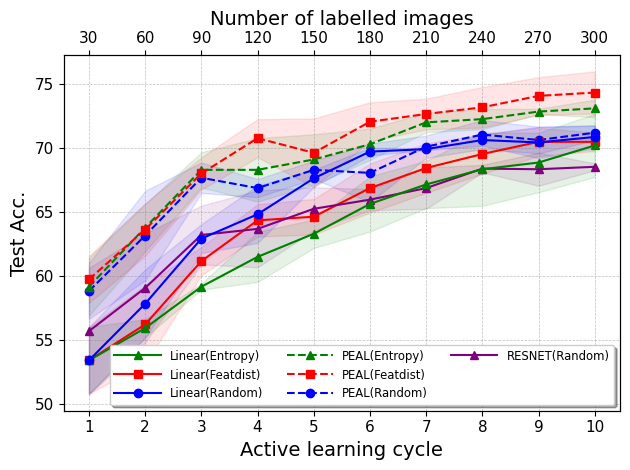}
    \caption{APTOS}
    \label{fig:aptos}
  \end{subfigure}
  \begin{subfigure}{0.33\linewidth}
\includegraphics[width=\textwidth, trim=0 0 0 0, clip]{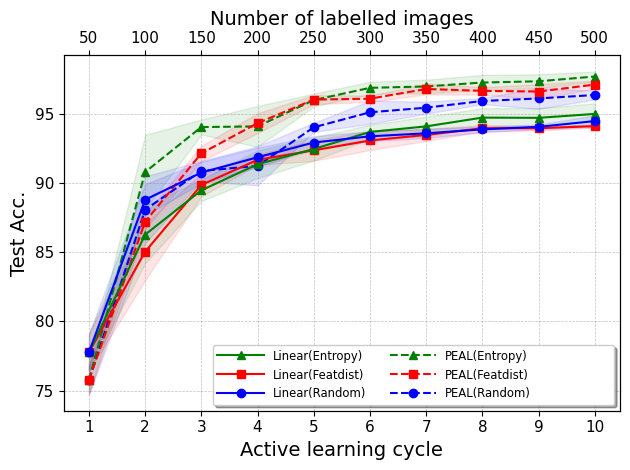}
    \caption{EuroSAT}
    \label{fig:eurosat}
  \end{subfigure}
\caption{Active learning with DINOv2 foundation model on various datasets. The plots compare accuracy as a function of amount of labeled data obtained from different sample selection strategies with linear probing versus PEAL.}
\label{fig:exp}
\end{figure*}

\subsection{Results}
Figure \ref{fig:exp} shows the results for the various datasets used for linear probing (solid-lines) and PEAL (dashed-lines) methods. As shown PEAL methods outperform linear methods. All experiments are subject to a transfer learning setup, where the model weights are not reinitialised to random weights in every AL cycle. This mirrors real-world settings where retraining from scratch each AL cycle is impractical. We report the average of $3$ trials over $10$ AL cycles.
 


\noindent \textbf{Histology:} Figure \ref{fig:histology} shows that using a ResNet-based random selection strategy results in suboptimal performance. Switching to a DINOv2 backbone with linear probing leads to an 8.5\% improvement initially, due to few-shot capabilities. Contrary to prior expectations in AL, linear probing methods using Entropy and Featdist sample selection underperform compared to random sampling. Employing parameter-efficient tuning methods like LoRA, \textit{PEAL (Random)} surpasses linear probing by 4\% in performance even at cycle 1. Furthermore, as sampling metrics get refined due to better feature representations and better calibrated entropy, \textit{PEAL~(Featdist)} and \textit{PEAL~(Entropy)} achieve 90\% accuracy with significantly fewer samples—250 and 200 respectively, compared to 400 for \textit{PEAL (Random)}. 

\noindent \textbf{APTOS:} Similar to Histology experiments, we observe that the PEAL based  methods outperform the Linear probing methods. 
The \textit{PEAL (Featdist)} method achieves 70\% test accuracy with only \textit{120} samples followed by \textit{180} samples by \textit{PEAL~(Entropy)}.

\noindent \textbf{EuroSAT:} The PEAL methods in Eurosat dataset achieve 93\% accuracy with just 200 samples as opposed to \textit{PEAL (Random)} which requires 400 samples. The superior performance of DINOv2 compared to ResNet backbone, as shown in Table \ref{tab:resnetv2eurosat}, highlights the few-shot capabilities of foundation models \cite{dinov2,clip,sam}. 

The superior performance of PEAL approach can be justified from Figure \ref{fig:histplots}, which compares linear probing and PEAL based approach on the ratio of unknown to known sample selection (where a higher ratio is preferable) during the initial cycles of active learning. An effective selection strategy would prioritize a greater number of incorrect samples (model unknowns) over correct samples (model knowns) throughout the AL cycles to maximize the efficacy of data annotation and model learning.


 \setlength{\textfloatsep}{1pt }
 \setlength{\abovecaptionskip}{1pt} 
 \setlength{\belowcaptionskip}{1pt} 
\begin{table}[]
\scalebox{0.9}{
\begin{tabular}{|c|c|c|}
\hline
\textbf{AL cycle} & \textbf{Resnet(Random}) & \textbf{PEAL(Random)} \\ \hline
1        & 47.23              & 75.75              \\ \hline
5        & 64.91              & 94.01              \\ \hline
10       & 69.6              & 95.9              \\ \hline
\end{tabular}
}
\centering
\caption{Test accuracy scores showing effect of model updates on AL cycles for EuroSAT dataset.}
\label{tab:resnetv2eurosat}
\end{table}
\textbf{Ablation study:} We explore class-balanced and class-agnostic sampling to evaluate sampling bias effects, as detailed in Table~\ref{tab:class-balanced}. For class-balanced sampling, we select $N_c = B/K$ samples per class based on the highest uncertainty or diversity metrics, where $B$ is the total budget and $K$ is the number of classes. Samples are labeled by the model, ensuring each class is equally represented in subsequent active learning cycles. In class-agnostic sampling, we simply choose the top-N samples irrespective of class. Our results show that class-balanced sampling markedly improves \textit{PEAL~(Featdist)} performance while \textit{PEAL~(Entropy)} sees less benefit, highlighting the need for tailored approaches in sampling strategy design.
\begin{table}[h]
\scalebox{0.71}{%
\begin{tabular}{@{}cccccccccc@{}}
\toprule
 && \multicolumn{2}{c}{\textbf{Histology}} && \multicolumn{2}{c}{\textbf{APTOS}} && \multicolumn{2}{c}{\textbf{EuroSAT}} \\ \cline{3-4}\cline{6-7}\cline{9-10}
AL Cycle && Entropy & Featdist && Entropy & Featdist && Entropy & Featdist \\ \midrule
1        && +3.00\%    & +11.00\%  && +0.20\%    & +5.00\%  && +2.00\%    & +3.10\%  \\ 
5        && +1.10\%    & +10.2\%  && +1.00\%    & +0.90\%  && +2.50\%    & +3.00\%  \\ 
10       && +0.70\%    & +2.00\%  && +2.10\%    & +3.80\%  && +1.67\%    & +2.78\%  \\ \bottomrule
\end{tabular}
}
\caption{Effect of class balanced sampling using PEAL techniques on different datasets. Values showcase the $\Delta$ improvements of class balanced sampling over class agnostic sampling. }
\label{tab:class-balanced}
\end{table}
\section{Conclusion}
In this study, we demonstrated parameter-efficient fine-tuning techniques within active learning. Our benchmarks show that our sampling strategies outperform linear probing baselines. These findings lay a foundation for advanced feature-based methods. While our diversity and uncertainty-based methods perform well across datasets, a hybrid strategy could enhance future results. Further exploration is needed to understand the impacts on tasks like semantic segmentation and object detection, which require extensive annotations. Addressing these challenges could improve model effectiveness in handling complex, real-world datasets.
{
    \small
    \bibliographystyle{ieeenat_fullname}
    \bibliography{main}
}


\end{document}